# Evaluating ChatGPT on Nuclear Domain-Specific Data

**Muhammad Anwar [1,2], Mischa de Costa [1,2], Issam Hammad [2] and Daniel Lau [1,2]**

[1] Enterprise Digital Technology, Digital Technology and Services, Ontario Power Generation, Pickering, Ontario, Canada

[2] Department of Engineering Mathematics and Internetworking, Faculty of Engineering, Dalhousie University, Halifax, Nova Scotia

muhammad.anwar@opg.com, mishca.decosta@opg.com, issam.hammad@dal.ca , daniel.lau@opg.com

## Abstract

This paper examines the application of ChatGPT, a large language model (LLM), for question-and-answer (Q&A) tasks in the highly specialized field of nuclear data. The primary focus is on evaluating ChatGPT's performance on a curated test dataset, comparing the outcomes of a standalone LLM with those generated through a Retrieval Augmented Generation (RAG) approach. LLMs, despite their recent advancements, are prone to generating incorrect or 'hallucinated' information, which is a significant limitation in applications requiring high accuracy and reliability. This study explores the potential of utilizing RAG in LLMs, a method that integrates external knowledge bases and sophisticated retrieval techniques to enhance the accuracy and relevance of generated outputs. In this context, the paper evaluates ChatGPT's ability to answer domain-specific questions, employing two methodologies: A) direct response from the LLM, and B) response from the LLM within a RAG framework. The effectiveness of these methods is assessed through a dual mechanism of human and LLM evaluation, scoring the responses for correctness and other metrics. The findings underscore the improvement in performance when incorporating a RAG pipeline in an LLM, particularly in generating more accurate and contextually appropriate responses for nuclear domain-specific queries. Additionally, the paper highlights alternative approaches to further refine and improve the quality of answers in such specialized domains.

## 1. Introduction

The nuclear industry is an OPEX (Operating Experience) driven industry. The sheer volume of accumulated OPEX data, spread across multiple formats and locations, makes finding relevant information a time-consuming challenge. OPG sought to utilize a chatbot for users to find answers to their nuclear domain-specific questions more easily.

To achieve this task, OPG turned to large language models (LLMs). LLMs have revolutionized Natural Language Processing (NLP) with their remarkable abilities to understand and produce language, impacting various artificial intelligence domains. These LLMs take the form of proprietary models offered as an API service such as ChatGPT, Cohere or Claude which are ready to use out of the box for general purpose tasks, as well as open-source model weights like Mistral and Llama which often require fine tuning before they can be used.

An emerging field in the NLP domain and LLM application development is Retrieval-augmented generation. The idea of RAG is that providing LLM high-quality facts from existing knowledge bases can enhance accuracy and context relevance. This could be especially useful in the nuclear domain where there is several decades of knowledge that is stored in various databases and document repositories. Providing a means for LLM to use that information to answer a question could greatly enhance their performance in nuclear domain.

Using LLMs greatly improves the user experience over traditional chatbots which relied on basic language models to detect "utterances" and then route the user through a series of prewritten messages based on the user's intent. LLMs on the other hand can provide much better understanding and generate natural text responses to users' questions. However, LLMs struggle with a few problems:

1. LLMs are expensive to train; training a foundational model is extremely costly, in the range of millions of dollars for a multibillion parameter model like ChatGPT. Without training on nuclear data, commercial and open source LLMs have a limited knowledge of the nuclear domain except for what little nuclear information from the public domain which is insufficient for day-to-day tasks in a nuclear power plant.
2. As a workaround to the above problem, LLMs can be provided information at inference time along with the users question as part of the prompt. However, this leads to the next problem which is that LLMs have a limited context window. In the early days of ChatGPT, this was limited to only 4096 tokens which is roughly 2-3 pages of text, hardly enough space to fit decades of nuclear information. The latest state of the art LLMs have massive context windows of over 100k tokens, but this is still insufficient to fit the vast amounts of data available, and other issues arise where LLMs forget information in the beginning part of long prompts.
3. The final concern is around hallucinations; as a side effect of RLHF (Reinforcement Learning from Human Feedback) where LLMs will make up incorrect facts in a very convincing manner. This behavior comes about from the LLM trying to produce a response a human would prefer, and humans do not like to be told "I don't know". Various techniques are available to suppress this behavior as will be discussed in this paper.

One of the emerging methods to address all three concerns above is to utilize Retrieval Augmented Generation (RAG) where data is semantically queried at runtime so that only the needed context can be passed to the model, and the model can be instructed to answer based on that information alone. The approach is discussed in the next section.

This paper primarily focuses on evaluating the performance of LLMs (specifically ChatGPT) when Retrieval Augmented Generation (RAG) is applied. The goal is to compare the quality of responses generated using RAG against direct responses from ChatGPT to determine if RAG can improve accuracy in the nuclear domain. This evaluation aims to provide strong evidence supporting the use of RAG for developing chatbots capable of generating accurate, nuclear-specific responses.

## 2.      Retrieval-Augmented Generation (RAG): A Background

Retrieval-Augmented Generation (RAG) is a technique that enhances the capabilities of large language models (LLMs) by grounding their responses in external knowledge sources. This approach is particularly valuable when LLMs need to operate in specialized domains or require access to rapidly changing information that may not be reflected in their pre-training data.

**Key Steps in the RAG Methodology**

1. **User Question:** The process begins with a user's query expressed in natural language.

2. **Query Expansion (Optional):** To optimize the search process, the query may be expanded to include synonyms, related terms, or disambiguate acronyms and technical jargon. This step can be performed using rule-based methods or fine-tuned language models.

3. **Information Retrieval:** RAG uses a combination of semantic and keyword-based search techniques to retrieve the most relevant documents from a knowledge base.

    - **Vector Search:** Vector databases containing pre-computed text embeddings enable semantic similarity comparisons. Cosine similarity is commonly used to identify the most relevant documents to the user's query.
    - **Keyword Search:** Techniques like BM25 help ensure that documents containing specific keywords are considered, boosting recall.
    - **Result Merging:** Search results from vector and keyword searches are combined using methods like Reciprocal Rank Fusion (RRF) to produce a final ranked list of documents.

4. **Augmented Prompt Construction:** The retrieved documents are formatted into a structured context, along with the user's query. This augmented prompt, along with domain-specific instructions, is provided as input to the LLM.

5. **Bot Response:** The LLM, guided by the context and instructions, generates a response that is both accurate and consistent with the provided knowledge base. The LLM should indicate when it is unable to find relevant information within the context.

**Advantages of RAG**

- **Access to Current Information:** RAG enables LLMs to stay up-to-date even in domains with rapidly changing knowledge.

- **Domain Specialization:** RAG allows LLMs to become domain experts, improving accuracy within specific fields.

- **Explainability:** By grounding responses in retrieved documents, RAG provides transparency into the LLM's reasoning process.

## 3.     Literature Review

To address the challenges of domain-specific question answering that leverages the power of LLMs in the nuclear domain, it's crucial to examine emerging research on domain-specific question-answering using these models and their potential applications in the nuclear industry.

Previous research has explored the potential of AI in enhancing safety and operational efficiency in CANDU-type nuclear power plants. Budzinski [1] investigates the application of machine learning techniques for verifying refueling activities, with implications for nuclear safeguards. Ahsan and Hassan [2] propose a machine learning-based system for predicting faults in the primary heat transport system of CANDU reactors, demonstrating the potential for proactive identification and prevention of system failures. Hammad et al. [3-4] employ deep learning and linear regression to automate the detection of flaws in ultrasonic scans of nuclear fuel channels, significantly improving the efficiency of safety inspections. Wallace et al. [5] utilize neural networks and ancillary data sources to enhance the localization of fuel defects, while their subsequent work [6] presents an ultrasonic inspection analysis tool that aids decision-making in defect identification. Collectively, these studies highlight the promising role of AI and machine learning in improving safety, efficiency, and decision-making across various aspects of nuclear power plant operations.

Research in this area also highlights the potential for enhancing Large Language Models (LLMs) to better handle industry-specific questions. One major obstacle is their limited understanding of specialized industry terminology and concepts [7]. To address this, researchers propose using smaller, domain-specific language models that act as "experts" to supplement the broader knowledge of LLMs [7]. Additionally, integrating LLMs with domain-specific question answering systems often incurs high costs, especially when large datasets of contextual information are required [8]. This highlights the need for cost-effective strategies, such as selectively reducing context size or employing less expensive LLMs for certain tasks [8]. These insights underscore how adapting LLMs for technical industries like the nuclear domain necessitates both accuracy and cost-efficiency considerations.

The issue of accuracy and reliability in LLM outputs becomes especially critical in the nuclear domain, where factual errors can have serious consequences. A recent survey delves into this complex problem, exploring how LLMs store and process knowledge to identify the root causes of inaccuracies [9]. Furthermore, research on improving large language models through external knowledge and automated feedback mechanisms demonstrates the potential of refining LLM responses to ensure factual correctness in domain-specific question answering [10]. These research efforts highlight the ongoing development of methods to enhance LLM trustworthiness, a crucial aspect for their successful integration within the nuclear industry.

It's important to be aware that LLMs tend to "hallucinate" – generating text that may seem plausible but contradicts established knowledge [11]. Understanding the various forms of hallucinations and techniques for grounding these models can help mitigate this issue in domain-specific scenarios. Continued research into better evaluation metrics, methods to increase controllability of LLM output, and ways to improve their explainability are vital to address the challenge of hallucinations. These advancements will further bolster the reliability of LLMs for domain-specific data within the nuclear sector.

In the domain of Natural Language Processing, Kant et al. [12] employ a Transformer model for sentiment classification. Although the model achieved a moderate F1 score of 0.69, the training process demanded substantial computational resources and time. Fine-tuning the model required intricate balancing and additional strategies, such as active learning, to address nuances in labels and class imbalances. Zhao et al. [13] introduce an approach to interpret CNN-based text classification using SHAP values, offering insights into model decisions at the cost of increased complexity. Abburi et al. [14] investigate the use of ensemble methods with various pre-trained LLMs, including BERT and its variants, for text classification tasks. While these models did not achieve top results in binary classification, they showed potential for nuanced classification and model attribution. The reviewed works suggest that LLMs, particularly when used in ensemble configurations, can more effectively capture the subtleties of language compared to traditional machine learning models like CNNs and LSTMs, which require significant training effort and are sensitive to data imbalances and nuances. Furthermore, using foundational models like GPT-4 can mitigate the need for extensive computational resources and time, making LLMs a compelling choice for complex text classification tasks.

## 4. Evaluation Methodology

This paper evaluates and compares ChatGPT's direct responses with those generated using Retrieval Augmented Generation (RAG), specifically within the nuclear domain. The goal is to determine which approach yields the most accurate and reliable results for LLM applications in the nuclear industry.

### *4.1 Evaluation Dataset*

The first step in this evaluation was curating a dataset specifically tailored for the nuclear domain. Since ChatGPT has been trained on most of the data available on the internet in the public domain, it was decided to select nuclear domain specific material which is publicly available. The Essential CANDU [15] textbook was selected for this purpose. This comprehensive resource is widely used for understanding CANDU nuclear reactors and the fundamentals of nuclear science and engineering within that context. The textbook's structure, split into chapters like Reactor Statics, Reactor Dynamics, Instrumentation and Control Systems, and Electrical Systems, provided a natural framework for our dataset.

To create a curated dataset, the textbook was carefully reviewed extracting relevant question-and-answer pairs. These answers serve as the 'ground truth' for evaluating ChatGPT's responses. Here are a few examples of these question-and-answer pairs:

**Question**: How is magnetic field created inside a synchronous generator in a CANDU Nuclear Power Plant?

**Answer (Ground Truth):** To create a magnetic field inside a synchronous generator, separate windings and an electrical power source must be used. This part of the generation system is known as the excitation system. The excitation system is essentially a controllable DC source. By adjusting the excitation system output voltage, the output voltage level of the generator can be controlled, and hence the reactive power output.

**Question:** What are the different types of generators in a CANDU Nuclear Power Plant?

**Answer (Ground Truth):** The main generator, the standby generators, and the generators in the emergency power system.

**Question:** In the context of CANDU Nuclear Power Plant, how long can class 3 power be interrupted for?

**Answer (Ground Truth):** Class III power can be interrupted for up to 5 minutes.

### *4.2  Answer Generation without RAG*

ChatGPT-3.5 offers significant advantages for question answering. Its training on a vast dataset provides extensive knowledge and the capacity to comprehend various question forms. Additionally, its generative nature enables it to craft responses that are both fluent and relevant, closely resembling human-written text. However, it's crucial to remain aware of two potential drawbacks: inherent biases and the risk of factual inaccuracies. Due to its training on a massive internet dataset, ChatGPT-3.5 may have internalized incorrect or misleading information.

Here's the input prompt provided to ChatGPT-3.5, designed to mitigate these risks:

```
Answer the question below.
Please stick to facts and avoid including any information which you're not sure about.

Question: {question}
```

The {question} placeholder gets populated iteratively with each question from the evaluation set.

### *4.3  Retrieval Augmented Generation*

To generate RAG-based responses for evaluation, a dedicated pipeline was developed. As outlined previously, a standard RAG pipeline consists of these key stages:

#### 4.3.1  Vector Index Creation

The 'Retrieval' aspect of RAG begins by dividing source documents into smaller 'chunks' to ensure consistent semantic meaning within each text segment. This enhances the quality of the vector embeddings generated for each chunk. We used OpenAI's ada-002 embedding model for this process. These embeddings were then incorporated into a Facebook AI Similarity Search (FAISS) vector index, enabling efficient retrieval during the query phase.

#### 4.3.2  Search and Prompt Augmentation

When a question is posed, the vector index is searched for relevant chunks that could aid in answering it. This search can be conducted using various methods; we opted for cosine similarity calculations between the query vector and those within the index.

For our evaluation, Facebook AI Similarity Search (FAISS) facilitated fast and scalable similarity searches. FAISS optimizes its speed through specialized indexing techniques that organize vectors for rapid identification of a query vector's nearest neighbors.

The most relevant retrieved passages are then combined with the original query as the input prompt for the LLM. This augmentation provides context and increases the probability of a factually grounded response. We used ChatGPT-3.5 for consistency with the non-RAG evaluation.

The following prompt structure is used for RAG:

```
Answer the question based only on the following context:
{context}
Please do not use any external information.
Please include as much information as possible in your response from the provided
context.

Question: {question}
```

The {context} placeholder gets populated with retrieved chunks from similarity search. The {question} placeholder gets populated iteratively with each question from the evaluation set.

### 4.3.3  Response Generation

The final stage utilizes a generative language model. Again, ChatGPT-3.5 was selected for consistency. Its fluency, coherence, and ability to follow instructions enable it to synthesize responses that align with the retrieved documents from the RAG pipeline.

### 4.4  Generation of Answers

To compare the performance of RAG against direct ChatGPT-3.5 responses, we generated answers for the questions in our evaluation dataset using both methods. Below, we illustrate the differences between these approaches with a few examples:

**Question:** How is magnetic field created inside a synchronous generator in a CANDU Nuclear Power Plant?
**Answer with RAG:** In a synchronous generator in a CANDU Nuclear Power Plant, the magnetic field is created by energizing the rotor winding through the field excitation circuit. As the turbine rotates the rotor, a rotating magnetic field is generated. The excitation current is supplied to the rotor winding through slip rings. The rotating magnetic flux induces a potential in the stator winding, resulting in the creation of the magnetic field inside the generator.
**Answer without RAG:** In a CANDU Nuclear Power Plant, the synchronous generator's magnetic field is generated by an excitation system, consisting of field windings around the rotor. When DC power—supplied by a DC generator or a rectifier system—is applied to these windings, a magnetic field is created. The rotating rotor within this magnetic field induces an electric current in the stator windings, which then produces the electrical output for the grid.

**Question:** What are the different types of generators in a CANDU Nuclear Power Plant?
**Answer with RAG:** The different types of generators in a CANDU Nuclear Power Plant are the main generator, standby generators, and generators in the emergency power system.
**Answer without RAG:** In a CANDU nuclear power plant, electricity is generated by two main types of generators: the main generator, which converts mechanical energy from a steam turbine into high voltage electrical energy for the grid, and the auxiliary generator, a smaller diesel-powered unit that provides backup power for essential systems during outages or maintenance when the main generator is offline. Both are integral for the plant's continuous and reliable power supply.

**Question:** In the context of CANDU Nuclear Power Plant, how long can class 3 power be interrupted for?
**Answer with RAG:** In the context of a CANDU Nuclear Power Plant, Class III power can be interrupted for up to 5 minutes.
**Answer without RAG:** In the context of CANDU Nuclear Power Plant, class 3 power can typically be interrupted for up to 72 hours. Class 3 power refers to the power supply required for the safe and controlled shutdown of the reactor and maintaining the cooling systems. This duration allows for necessary maintenance, repairs, or other unforeseen events while ensuring the safety and stability of the plant. It is important to note that this duration may vary depending on plant-specific safety protocols and regulatory requirements.

## 4.5   Evaluation of Answers

To thoroughly assess the generated responses, we employed a multifaceted evaluation approach:

### 4.5.1   Human Evaluation

A human evaluator meticulously compared answers to the ground truth, establishing factual correctness. This provides a direct measure of accuracy.

### 4.5.2   LLM Based Evaluation

Recognizing the potential for scalability and interpretability, we leveraged cutting-edge LLMs as judges. Recent research demonstrates remarkable agreement between powerful LLMs like GPT-4 and human preferences [16]. This makes LLM-as-a-judge a viable alternative to potentially costly human evaluations.

For our experiment, GPT-4 assessed the responses generated with and without RAG. We provided the ground truth as context to enable informed evaluations. The GPT-4 evaluation prompt was meticulously crafted to guide judgments. Specific criteria was included for scoring responses, including helpfulness, grounded-ness, correctness, conciseness, coherence, and detail. GPT-4 generated scores for each criterion along with explanations. A final verdict and justification were requested, based on these scores. This structured prompt promotes thoughtful analysis and insightful outputs. The full evaluation prompt is shown below:

Please act as an impartial judge and evaluate the quality of the responses provided by two AI assistants to the user question displayed below.

You should choose the assistant that follows the user's instructions and answers the user's question better based on the ground truth.

For this evaluation, you should primarily consider the following criteria:

HELPFULNESS: "Is the submission helpful, insightful, and appropriate?"

GROUNDEDNESS: "Is the submission grounded in facts as compared to the ground truth provided?"

CORRECTNESS: "Is the submission correct, accurate, and factual?"

CONCISENESS: "Is the submission concise and to the point?"

COHERENCE: "Is the submission coherent, well-structured, and organized?"

DETAIL: "Does the submission demonstrate attention to detail?"

Your output will be as follows:

1. Begin your evaluation by comparing the two responses and provide a short explanation, then provide a criterion score for each the two responses as a number from 0 (lowest) to 100 (highest) with granular integral intervals, in output format:
"CRITERION
Evaluation: explanation.
Scores: A: score, B: score"

Avoid any position biases and ensure that the order in which the responses were presented does not influence your decision. Avoid allowing the length of the responses to influence your evaluation. Avoid favouring certain names of the assistants. Be as objective as possible.

2. After providing your criteria explanation and scores, provide a summary explanation of your final verdict as:
"Verdict Evaluation: evaluation"

3. Then assuming criteria are weighted equally calculate the arithmetic mean for each response using the formula sum (scores for each criterion) / number of criteria, and strictly output in the following format filling in score:
"Verdict Scores: A: score, B: score"

```
4. After providing your explanation and scores, output your final verdict as:
"Verdict Preference: "
then strictly following this format:
[[A]] if assistant A is better, [[B]] if assistant B is better, and [[C]] for a
tie.

Double-check the explanations, scores, and final verdicts have been included.
```

As seen above in the prompt, instructions were also provided in the prompt to ensure the output from GPT-4 follows a consistent pattern and therefore can be parsed out in a reliable way. This is important otherwise it will require manual effort to collate all the LLM responses to make them presentable.

## 5. Results

Our evaluation demonstrates the clear superiority of responses generated using the RAG pipeline compared to direct ChatGPT-3.5 output. Overall verdict scores for RAG responses were significantly higher.

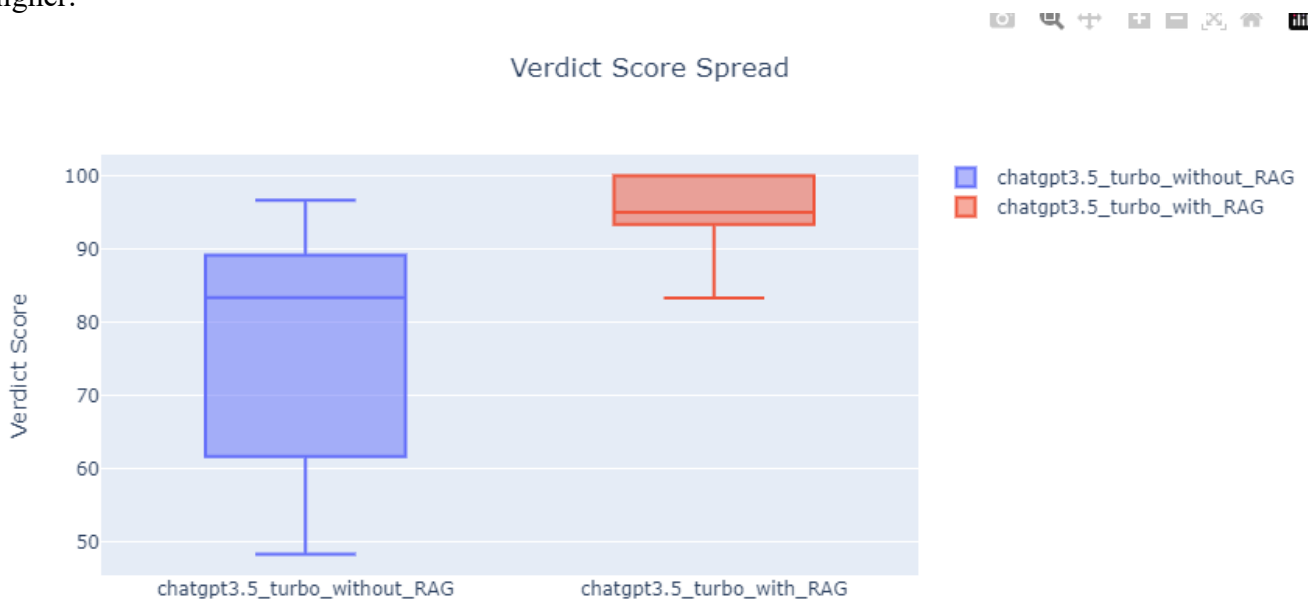

Figure 1: Box Plot comparing responses for ChatGPT3.5 with RAG and without RAG

This above box plot visually underscores the consistent improvement with RAG. It also reinforces the findings of our human evaluator, who consistently rated RAG responses as equal or superior to non-RAG responses.

Furthermore, RAG responses exhibit remarkable consistency, with scores tightly clustered between 90 and 100. Conversely, direct ChatGPT-3.5 responses demonstrate greater variability (60 to 90), highlighting the inherent randomness sometimes found in its output. This suggests that the RAG pipeline provides crucial context, anchoring LLM responses and enhancing their accuracy.

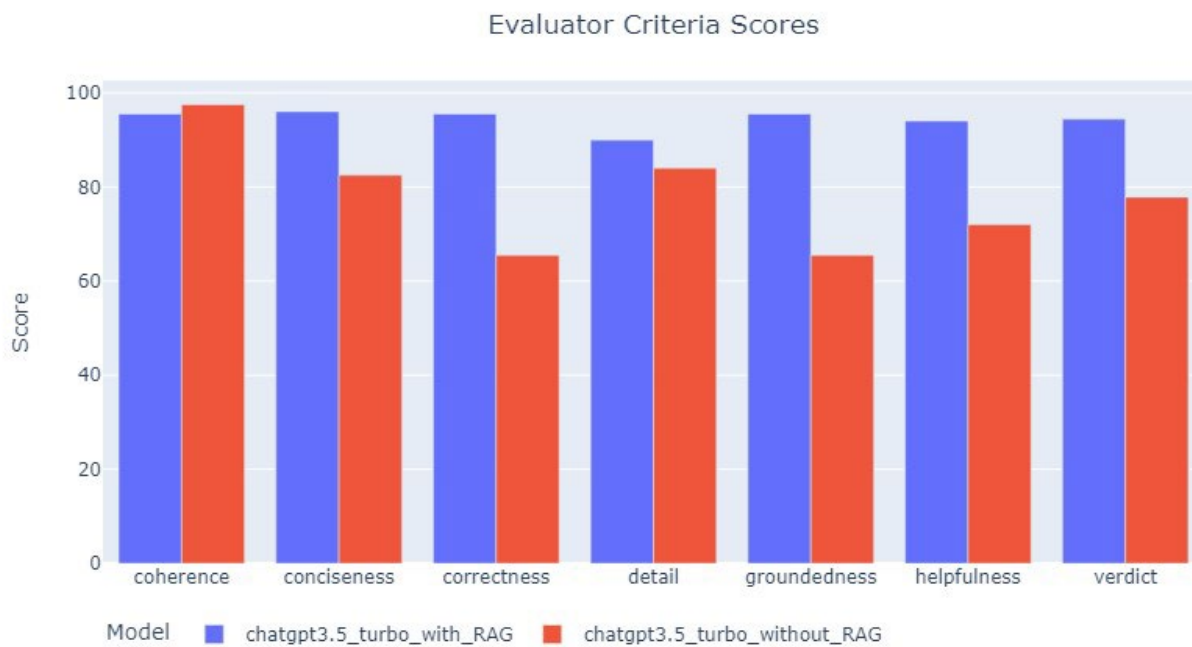

Figure 2: Column Plot comparing scores for ChatGPT3.5 with RAG and without RAG

This above plot offers a direct comparison of average metric scores. RAG-generated responses significantly outperform direct ChatGPT-3.5 responses in correctness, grounded-ness, and helpfulness – all key indicators of factual accuracy and alignment with the ground truth.

Additionally, RAG-augmented responses surpass their counterparts in conciseness. This aligns with the expectation that LLMs provided with targeted context and explicit instructions gravitate towards more focused outputs. Coherence and detail scores show no significant difference between the two approaches.

## 6. Conclusion and Future Work

Our evaluation convincingly demonstrates the value of a RAG pipeline in ensuring the factual accuracy, relevance, helpfulness, and conciseness of LLM-generated responses. This aligns with the fundamental architecture of LLMs, which rely on probabilistic predictions based on their training data. While their vast language knowledge is impressive, the lack of curated training datasets in specific domains can lead to factual inaccuracies.

By equipping LLMs with RAG pipeline, we provide targeted context and reinforce their ability to follow instructions, leading to significantly improved results. This work can be further improved through improved Information retrieval and fine-tuning LLMs on nuclear data as illustrated in Sections 6.1 and 6.2.

## 6.1 Improved Information Retrieval

The RAG pipeline can be improved by improving the information retrieval step. This can be done using the following techniques:

- **Context Aware Chunking:** During information retrieval, the vector index is searched for similar chunks based on similarity score between the vector embedding of the question and all the chunks from the document. Hence the search performance is directly related to the quality of vector embeddings. To obtain better quality vector embeddings, context aware chunking techniques can be used which can ensure that the semantic context within a chunk is uniform.

- **Expansion of acronyms and technical jargon:** Since the vector embeddings are a numeric representation of the context contained within input text, the quality of embeddings can be enhanced by expanding the acronym or technical jargon included in the text. This is because the embeddings model is likely to have seen the acronyms/technical jargon during its training.

## 6.2 Fine-tuning LLM on Nuclear Data

Large language models (LLMs) have a vast understanding of general language, but often need refinement for specific applications. Fine-tuning involves training an LLM on a smaller, domain-specific dataset. This adapts the model's weights to better understand the terminology, nuances, and style of that particular domain (e.g., legal, financial or nuclear). Fine-tuning significantly improves LLM performance within these specialized areas.

While full fine-tuning can be computationally expensive, Parameter-Efficient Fine-Tuning (PEFT) offers a solution. PEFT selectively updates only a small portion of the LLM's parameters, either by adding task-specific layers or modifying a subset of existing ones. This approach dramatically reduces computational and storage requirements while maintaining performance levels close to full fine-tuning

## 7. Acknowledgements

This research was funded by The Natural Sciences and Engineering Research Council of Canada (NSERC) and the Canadian Nuclear Safety Commission (CNSC).